# A Sensitivity Analysis of Pathfinder: A Follow-up Study

**Keung-Chi Ng**
Department of Pathology
University of Southern California
HMR 204, 2025 Zonal Ave
Los Angeles, CA 90033

**Bruce Abramson**
Department of Computer Science
University of Southern California
Los Angeles, CA 90089-0782

## Abstract

At last year's Uncertainty in AI Conference, we reported the results of a sensitivity analysis study of Pathfinder. Our findings were quite unexpected—slight variations to Pathfinder's parameters appeared to lead to substantial degradations in system performance. A careful look at our first analysis, together with the valuable feedback provided by the participants of last year's conference, led us to conduct a follow-up study. Our follow-up differs from our initial study in two ways: (i) the probabilities 0.0 and 1.0 remained unchanged, and (ii) the variations to the probabilities that are close to both ends (0.0 or 1.0) were less than the ones close to the middle (0.5). The results of the follow-up study look more reasonable—slight variations to Pathfinder's parameters now have little effect on its performance. Taken together, these two sets of results suggest a viable extension of a common decision analytic sensitivity analysis to the larger, more complex settings generally encountered in artificial intelligence.

## 1 INTRODUCTION

A great deal of recent attention has been focused on the relationship of artificial intelligence (AI) to decision analysis (DA). Most of the work on probability theory as a mechanism for uncertainty management (Cheeseman 1988; Ng and Abramson 1990b), belief networks as representations of uncertainty (Abramson 1990; Howard and Matheson 1984), the propagation of information through a belief network (Pearl 1988; Shachter 1986; Shachter 1987), and the design of systems based on belief networks (Abramson and Finizza 1991; Andereassen, Woldbye, Falck, and Anderson 1987; Heckerman, Horvitz, and Nathwani 1990), falls into this category. These topics are all familiar to decision analysts; they deal with the representation of uncertainty and the derivation of inference. There is, however, a third concern frequently studied in DA: analysis.

In order to be useful, a model and/or system must pass through several stages of development: it must capture the information that it claims to be modeling, it must allow questions to be answered, and it must be (in some sense) validated. Sensitivity analyses fall into this last area of concern. In a classic DA sense, a sensitivity analysis measures the degree to which a decision is sensitive to changes in its inputs. These analyses are generally done one variable at a time. Most well-designed models exhibit a phenomenon known as a *flat maximum;* small (and even medium-sized) changes in input variables rarely lead to changed decisions (von Winterfeldt and Edwards 1986).

Sensitivity analyses are as important to AI systems as they are to DA models. Two characteristics of AI systems, however, force the standard techniques of sensitivity analyses to be rethought: they generate outputs other than decisions, and they include huge numbers of variables. Pathfinder, for example, is an AI system that diagnoses the 63 diseases of the human lymph system; its underlying belief network also contains over 100 distinct symptoms (Heckerman, Horvitz, and Nathwani 1990). In 1990, at the *Sixth Conference on Uncertainty in Artificial Intelligence,* we presented a sensitivity analysis of Pathfinder that led to a surprising conclusion: the system did not exhibit a flat maximum. Instead, it seemed to be highly sensitive to the parameters specified by its contributing expert; even minor changes to these parameters led to drastic declines in the system's performance (Ng and Abramson 1990a). These results were presented with a fairly detailed description of the modified sensitivity analysis techniques upon which our studies were based, and a challenge to the conference participants to suggest possible causes underlying our results. Several useful suggestions were made. This paper revises our modifications to the analysis and provides results that are more in line with previous (empirical) experience.

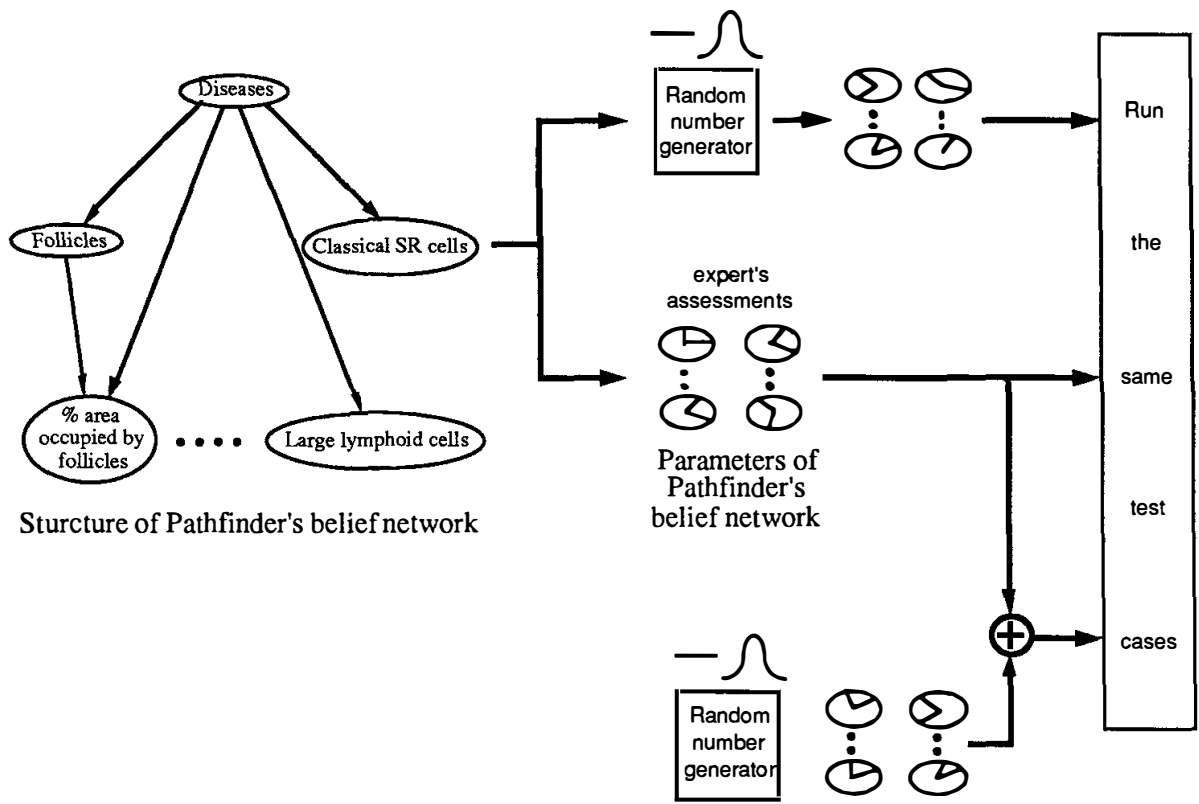

Figure 1: Block diagram of the sensitivity analysis of Pathfinder.

## 2 DETAILS OF THE ANALYSIS

The goals of a sensitivity analysis are (i) to gain insight into the nature of a problem, (ii) to find a simple and elegant structure that does justice to the problem, and (iii) to check the correctness of the numbers and the need for precision in refining them (von Winterfeldt and Edwards 1986). In most decision problems, once the numbers have reached a certain degree of precision, further refinement of these numbers has little effect on the decisions. Our studies are directed towards determining whether or not similar observations are true for diagnostic systems, (i.e., once the prior and conditional probabilities have reached certain quality, further improvement on these probabilities has little effect on its diagnoses).

A block diagram of the study is shown in Figure 1. In our study, experiments were run on a body of 60 "classic" cases in which the diagnosis was known. Since a network's parameters include prior and conditional probabilities, both sets of probabilities had to be varied. The experiments used two sets of prior probabilities (those specified by the experts and a uniform distribution across all the diseases) and three types of conditionals:

1. The original values, exactly as assessed by experts.

2. Randomly generated probabilities. This class of parameters includes probabilities distributed both uniformly and normally.

3. The values assessed by experts plus randomly generated noise, using both uniformly and normally distributed noise functions.

All the conditional probabilities, either generated or augmented with noise, were renormalized.

Each body of tests served a different purpose; the original knowledge base defined a standard against which others would be judged, the random parameters addressed the relative importance of parameters, and the random noise addressed the issue of sensitivity. The use of two different sets of priors addressed the effect of priors on system performance.

## 3 THE INITIAL STUDY

The results discussed in this section were first reported at the 1990 Uncertainty in AI Conference (Ng and Abramson 1990a). This study implied that Pathfinder's performance degraded so significantly with randomly generated probabilities that the resultant system had negligible discriminating power. The same results were observed regardless of the choice of distribution function or the selection of priors. These findings led us to conclude that parameters are crucial to a belief network (or at least to Pathfinder's belief network) and that experts are needed to provide the parameters.

Our experiments also studied variations in both prior and conditional probabilities. Priors were fixed either at the expert-assessed set or at a uniform set. Conditionals were varied by augmenting the expert's assessment with randomly generated noise. The resultant conditional probabilities were then renormalized. The random noise functions followed uniform or normal distributions with mean of 0 ($\mu = 0$) and several values of standard deviation ($\sigma$). For each noise function, five parameter sets were created. Sixty cases were run on each network, for a total of 300 test cases per noise function. A total of seven noise generating schemes were used, including uniformly generated noise (uniform noise), normally generated noise (normal noise) with $\sigma$ of 0.005, 0.01, 0.025, 0.05, 0.1, 0.25, all with $\mu = 0$ (to ensure that any probability has equal chance of being increased or decreased). A summary of the test results with expert priors is shown in Table 1. In the table, the percentage of correct diagnoses, (i.e., the number of cases in which the known diagnosis was assigned the highest posterior probability divided by the total number of cases), is intended to provide a measure of the diagnostic power. The average confidence, (average difference in posterior probabilities between the two diseases with the highest posterior probabilities on the differential diagnosis for all the cases run), with respect to the correct diagnosis and incorrect diagnosis,[1] provides a measure of the discriminating power of the leading disease (disease with the highest posterior probability on the differential diagnosis) from the other diseases on the differential diagnosis. It should be pointed out that although the percentage of correct diagnoses is more important than average confidence in system performance, average confidence is also useful in gaging system performance. Systems scoring perfectly (100%) in the correct diagnosis column with 0 average confidence (e.g., all diseases have the same posterior probability with respect to all the test cases) could be as useless a system as one with no correct diagnoses and absolute average confidence (1.0). Also shown in Table 1, in the column headed "Percentage Better," is the percentage of cases in which the noisy network assigned the correct diagnosis with posterior probability that is notably higher than did the original network.

Table 1 indicates that the original knowledge base had the highest score in both percentage of correct diagnoses and average confidence; augmentation with uniform noise produced the lowest scores on both items. Adding normal noise to the original knowledge base produced a system with scores that lie between these extremes, with better results for systems with smaller standard deviations (or less noise). Furthermore, the chance of producing a better diagnosis than the original knowledge base is higher for knowledge bases with less noise than those with more noise. Table 2 summarizes the results of networks with equal priors. The results are similar to those of Table 1.

## 4   THE FOLLOW-UP STUDY

The results summarized in Tables 1 and 2 implied that Pathfinder's belief network did not exhibit a flat maximum. This observation was quite unexpected; flat maxima have been observed in almost all tested models (von Winterfeldt and Edwards 1986). A closer look at our study revealed that one possible reason for this surprise: all probabilities were varied, including those that were equal to 0.0 and 1.0. The variation of these probabilities was undoubtedly a mistake; whereas probabilities in $(0, 1)$ represent degrees of belief that may be varied and refined, probabilities at the endpoints represent *absolute certainty*. Although studying perturbations of beliefs is reasonable, studying perturbations of certainty is not. Furthermore, beliefs close to 0.0 and 1.0 are less prone to adjustment than those located elsewhere.

In an attempt to account for this observation, we reran our study using a scheme that generated noise in a slightly different way; the augmented noise function depended on both the random noise function and the actual value of the conditional probability. If the original conditional probability was 0.0 or 1.0 (i.e., the relationship is crisp or definitional), the conditional probability remained unchanged. For other values of conditional probability with normally distributed noise functions, the noise would be more for probabilities close to 0.5 than for those close to either 0.0 or 1.0. The approach that we used is (i) convert the original probability $p$, with range (0.0, 1.0), to a value $v$ of range $(-\infty, \infty)$, by a function $f(p) = \ln(\frac{p}{1-p})$, (ii) add the generated noise to $v$ to obtain $v'$, and then (iii) convert $v'$ back to a probability by using $f^{-1}$ (the inverse of the function $f$). With this scheme, the changes in likelihood ratios due to the added noise are comparable for the probabilities that are close to the endpoints (0.0 and 1.0) and those that are close to 0.5, while the actual variations in probabilities are smaller for probabilities that are close to either ends than those near 0.5. The new results are shown in Tables 3 and 4. In these tables, the average improvement in posterior probability for the cases in which the noisy network outperforms the original network is also shown (under the column labelled "average amount better").

---

[1] Correct diagnosis is defined as the situation in which the disease with the highest posterior probability on the differential diagnosis provided by the system is the same as the known diagnosis for a test case; incorrect diagnosis is the situation in which the disease with the highest posterior probability on the differential diagnosis is different from the known diagnosis.

|  | Percentage Correct | Average Confidence | | Percentage Better |
|---|---|---|---|---|
|  |  | Correct (# cases) | Incorrect (# cases) |  |
| Original knowledge base | 98.3% | 0.7910 (59) | 0.8247 (1) | — |
| Normal noise, SD=0.005 | 90.0% | 0.7329 (270) | 0.2511 (30) | 22.0% |
| Normal noise, SD=0.01 | 87.6% | 0.6963 (263) | 0.3685 (37) | 21.6% |
| Normal noise, SD=0.025 | 78.0% | 0.6803 (234) | 0.2692 (66) | 16.3% |
| Normal noise, SD=0.05 | 70.7% | 0.6997 (212) | 0.2541 (88) | 15.7% |
| Normal noise, SD=0.1 | 61.0% | 0.6212 (183) | 0.3262 (117) | 9.3% |
| Normal noise, SD=0.25 | 36.7% | 0.5614 (110) | 0.2988 (190) | 6.0% |
| Uniform noise | 32.7% | 0.5417 (98) | 0.3142 (202) | 5.0% |

Table 1: Summary of results of the initial (1990) sensitivity analysis of Pathfinder. In this set, expert-assessed priors were used for all networks. A variety of noise functions were added to the expert's conditional probabilities.

|  | Percentage Correct | Average Confidence | | Percentage Better |
|---|---|---|---|---|
|  |  | Correct (# cases) | Incorrect (# cases) |  |
| Original knowledge base | 88.3% | 0.8262 (53) | 0.1569 (7) | — |
| Normal noise, SD=0.005 | 84.7% | 0.7900 (254) | 0.1279 (46) | 24.3% |
| Normal noise, SD=0.01 | 83.0% | 0.7688 (249) | 0.1444 (51) | 17.7% |
| Normal noise, SD=0.025 | 80.0% | 0.7019 (240) | 0.2052 (60) | 18.3% |
| Normal noise, SD=0.05 | 73.7% | 0.6784 (221) | 0.1831 (79) | 13.7% |
| Normal noise, SD=0.1 | 62.3% | 0.6458 (187) | 0.3019 (113) | 10.0% |
| Normal noise, SD=0.25 | 42.3% | 0.4914 (127) | 0.2516 (173) | 7.3% |
| Uniform noise | 36.7% | 0.4347 (110) | 0.2927 (190) | 6.0% |

Table 2: Summary of results of the initial (1990) sensitivity analysis of Pathfinder. In this set, uniform priors were used for all networks. A variety of noise functions were added to the expert's conditional probabilities.

| | Percentage Correct | Average Confidence | | Percentage Better (average amount better) |
|---|---|---|---|---|
| | | Correct (# cases) | Incorrect (# cases) | |
| Original knowledge base | 98.3% | 0.7910 (59) | 0.8247 (1) | — |
| Normal noise, $\sigma=0.005$ | 96.6% | 0.8060 (290) | 0.4161 (10) | 1.7% (0.0678) |
| Normal noise, $\sigma=0.01$ | 96.0% | 0.8119 (288) | 0.3615 (12) | 6.3% (0.0293) |
| Normal noise, $\sigma=0.025$ | 96.0% | 0.8101 (288) | 0.3624 (12) | 10.3% (0.0306) |
| Normal noise, $\sigma=0.05$ | 95.3% | 0.8130 (286) | 0.3931 (14) | 13.3% (0.0485) |
| Normal noise, $\sigma=0.1$ | 96.3% | 0.8090 (289) | 0.4573 (11) | 17.0% (0.0680) |
| Normal noise, $\sigma=0.25$ | 90.3% | 0.8270 (271) | 0.4204 (29) | 18.3% (0.1237) |
| Uniform noise | 65.3% | 0.8495 (196) | 0.4798 (104) | 11.0% (0.1989) |

Table 3: Summary of results of the sensitivity analysis of Pathfinder. In this set, expert-assessed priors were used for all networks. A variety of noise functions were added to the expert's conditional probabilities. In this study, more noise was added to probabilities that were close to 0.5 than those that were close to either 0.0 or 1.0, and no change would be made to probabilities that were 0.0 or 1.0.

Table 3 indicates that the percentage of correct diagnoses and average confidence for the original knowledge base and knowledge bases with small noise are comparable. With more noise added to the original knowledge base, the percentage of correct diagnoses drops, but the average confidence remains comparable to that of the original knowledge base. Knowledge bases with more noise, however, have a higher chance of providing better diagnoses and the improvements are more significant (see the column under percentage better)—an average improvement of more than 0.1 in posterior probability for the correct diagnosis over that of the original knowledge base, both for normal noise with $\sigma = 0.25$ and uniform noise. All these observations suggest that although small refinements to the probabilities will not produce significant differences in performance, larger proper refinements may produce stronger results (at least with respect to the test cases). Table 4 summarizes the results of networks with equal priors. The results are similar to those of Table 3.

The results in Table 1 to 4 contain a lot of fine details on our study, which can be nicely summarized through the use of a scoring rule. Table 5 shows the results of our earlier study and this follow-up study evaluated by a quadratic scoring rule (see von Winterfeldt and Edwards 1986 for examples of popular scoring schemes). The scoring rule that we used is

$$score = 2 * p_k - \sum_i (p_i)^2,$$

where $p_i$ denotes the probability assigned to hypothesis $i$, and $p_k$ is the probability assigned to the correct hypothesis $k$, which is the correct disease in a test case in our study. It can be observed that score can take on values between 1.0 and -1.0, with 1.0 denoting a perfect score—in Pathfinder, this denotes that the system's diagnosis is the same as the diagnosis in the test cases. The values shown in the table are the average scores of each study for the 300 test cases. The results shown in Table 5 revealed the same key message as in Tables 1 to 4, however, the fine details of our results can only be found in Tables 1 to 4.

## 5  CONCLUSIONS

The results presented in this paper are hardly revolutionary; they indicate that Pathfinder exhibits a flat maximum. They are, however, interesting in several respects. First, they help extend the flat maximum phenomenon beyond decision settings to diagnostic settings. Second, they show how to extend an analysis that is usually conducted one variable at a time to one that can be conducted on all of a domain's variables simultaneously. Third, they reveal the importance of parameters to a belief network, especially those that are close to (or at) 0.0 and 1.0. Fourth, (and of perhaps greatest significance to the authors), they correct results that we have already published. This paper, combined with our earlier one (Ng and Abramson 1990a), then, suggest a viable extension of a common DA analysis to the larger, more complex settings generally encountered in AI.

A Sensitivity Analysis of Pathfinder: A Follow-Up Study    247| | Percentage Correct | Average Confidence | | Percentage Better (average amount better) |
|---|---|---|---|---|
| | | Correct (# cases) | Incorrect (# cases) | |
| Original knowledge base | 88.3% | 0.8262 (53) | 0.1569 (7) | — |
| Normal noise, $\sigma=0.005$ | 86.0% | 0.8499 (258) | 0.1335 (42) | 3.3% (0.0280) |
| Normal noise, $\sigma=0.01$ | 86.3% | 0.8464 (259) | 0.1304 (41) | 6.3% (0.0238) |
| Normal noise, $\sigma=0.025$ | 85.7% | 0.8562 (257) | 0.1388 (43) | 12.0% (0.0312) |
| Normal noise, $\sigma=0.05$ | 86.7% | 0.8444 (260) | 0.1933 (40) | 14.7% (0.0398) |
| Normal noise, $\sigma=0.1$ | 85.7% | 0.8583 (257) | 0.2187 (43) | 19.7% (0.0616) |
| Normal noise, $\sigma=0.25$ | 84.0% | 0.8331 (252) | 0.3793 (48) | 16.7% (0.1274) |
| Uniform noise | 66.7% | 0.8707 (200) | 0.4628 (100) | 12.0% (0.2145) |

Table 4: Summary of results of the sensitivity analysis of Pathfinder. In this set, uniform priors were used for all networks. A variety of noise functions were added to the expert's conditional probabilities. In this study, more noise was added to probabilities that were close to 0.5 than those that were close to either 0.0 or 1.0, and no change would be made to probabilities that were 0.0 or 1.0.

| | First Study | | Follow-up Study | |
|---|---|---|---|---|
| | Expert Priors | Equal Priors | Expert Priors | Equal Priors |
| Original knowledge base | 0.8829 | 0.8393 | 0.8829 | 0.8393 |
| Normal noise, $\sigma=0.005$ | 0.8151 | 0.8114 | 0.8837 | 0.8392 |
| Normal noise, $\sigma=0.01$ | 0.7542 | 0.7891 | 0.8835 | 0.8399 |
| Normal noise, $\sigma=0.025$ | 0.6695 | 0.7052 | 0.8824 | 0.8390 |
| Normal noise, $\sigma=0.05$ | 0.6045 | 0.6619 | 0.8774 | 0.8333 |
| Normal noise, $\sigma=0.1$ | 0.4408 | 0.4727 | 0.8809 | 0.8276 |
| Normal noise, $\sigma=0.25$ | 0.1480 | 0.2226 | 0.8255 | 0.7660 |
| Uniform noise | 0.0748 | 0.1055 | 0.4923 | 0.5186 |

Table 5: A table summarizing our first study and this follow-up study of the sensitivity analysis of Pathfinder. The number in each entry denotes the average score computed by a quadratic scoring rule.

## Acknowledgments

This work was supported in part by the National Library of Medicine under grant R01LM04529 and by the NSF under grant IRI-8910173.